\pdfoutput=1

\documentclass[11pt]{article}

\usepackage[preprint]{acl}

\usepackage{times}
\usepackage{latexsym}

\usepackage[T1]{fontenc}

\usepackage[utf8]{inputenc}

\usepackage{microtype}

\usepackage{inconsolata}

\usepackage{graphicx}

\title{Exploring Adversarial Robustness in Classification tasks \\ using DNA Language Models}

\author{
  Hyunwoo Yoo$^1$ \quad \quad
  Haebin Shin$^2$ \quad \quad
  Kaidi Xu$^1$ \quad \quad
  Gail Rosen$^1$ \quad \quad
  \\
  \\
  $^1$Drexel University \quad $^2$KAIST AI
  \\[2pt]
  \texttt{\{hty23, kx46, glr26\}@drexel.edu} \quad \texttt{haebin.shin@kaist.ac.kr}
}

\begin{document}
\maketitle
\begin{abstract}

DNA Language Models, such as GROVER, DNABERT2 and the Nucleotide Transformer, operate on DNA sequences that inherently contain sequencing errors, mutations, and laboratory-induced noise, which may significantly impact model performance.
Despite the importance of this issue, the robustness of DNA language models remains largely underexplored.
In this paper, we comprehensivly investigate their robustness in DNA classification by applying various adversarial attack strategies: the character (nucleotide substitutions), word (codon modifications), and sentence levels (back-translation-based transformations) to systematically analyze model vulnerabilities.
Our results demonstrate that DNA language models are highly susceptible to adversarial attacks, leading to significant performance degradation. 
Furthermore, we explore adversarial training method as a defense mechanism, which enhances both robustness and classification accuracy.
This study highlights the limitations of DNA language models and underscores the necessity of robustness in bioinformatics.

\end{abstract}

\section{Introduction}

Transformer-based language models are increasingly being adopted in bioinformatics, leveraging NLP techniques to tackle sequence classification and functional prediction tasks, which have traditionally relied on alignment-based methods\cite{steinegger2017mmseqs2, buchfink2021diamond}.
Notably, language models such as DNABERT2~\cite{zhou2023dnabert}, Nucleotide Transformer \cite{dalla-torre2023nucleotide}, and GROVER \cite{sanabria2024grover} leverage large-scale genomic sequence data as textual data and are specialized for specific bioinformatics downstream tasks. By treating DNA sequences not merely as strings but as sequence data with contextual information, these models introduce a novel approach to solving bioinformatics problems.



\begin{figure}[h]
    \centering
    \includegraphics[width=0.95\linewidth]{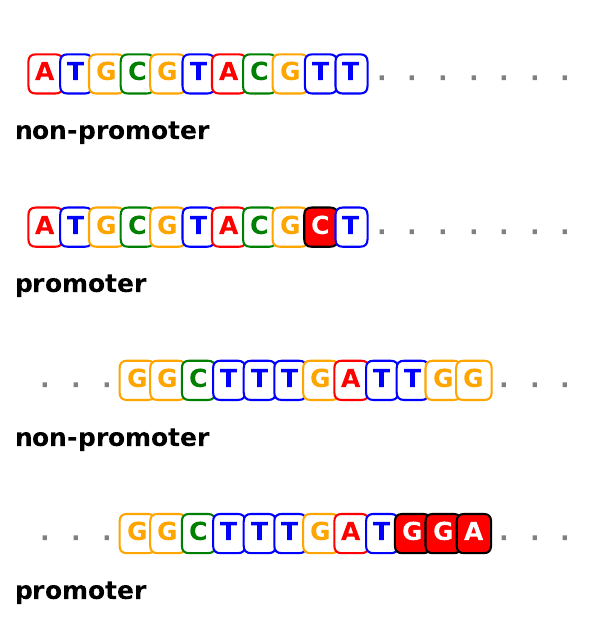}
    \caption{Adversarial examples with nucleotides and codon change,
which will be misclassified by a non-promoter to a promoter.}
    \label{fig:dna_attack_visualization}
\end{figure}

However, despite these advancements, the robustness of DNA language models remains underexplored. In real-world biological environments, DNA sequences are susceptible to sequencing errors, mutations, and data noise introduced during the extraction process in laboratories \cite{Ono2021PBSIM2, Ma2019ErrorProfiles}. However, there is a lack of systematic research analyzing the impact of such variations on model performance. While robustness studies on language models have been active in text classification, research on DNA sequence classification models remains limited. Given the increasing adoption of DNA classification models in clinical and biotechnological applications, it is crucial to assess their reliability in real-world scenarios.

\begin{figure*}
    \centering
    \includegraphics[width=0.95\linewidth]{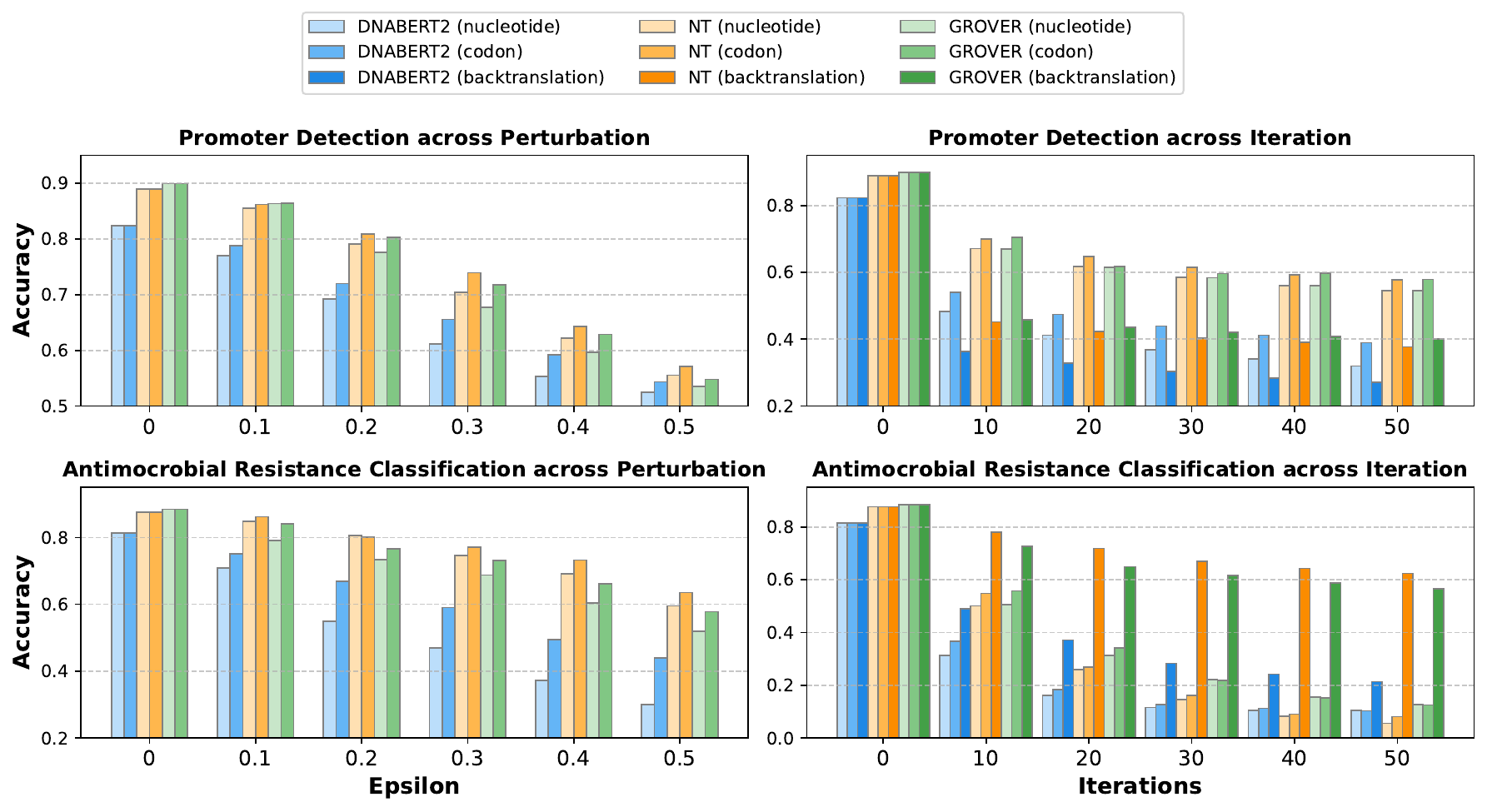}
    \caption{Anti-Microbial Resistance (AMR) gene Classification and Promoter Detection Results}
    \label{fig:pd_amr_accuracy}
\end{figure*}

In this paper, we explore the adversarial robustness of DNA language models in classification tasks. In addition to existing character-level attacks (nucleotide-level attacks \cite{kuleshov2021quantifying}), we introduce word-level (codon-level) and sentence-level (backtranslation) attacks. We demonstrate that these attacks lead to significant performance degradation in Anti-Microbial Resistance (AMR) gene classification~\cite{yoo2024predicting} and promoter detection~\cite{zhou2023dnabert}. 

Furthermore, we investigate whether a simple adversarial training strategy can mitigate the vulnerabilities of DNA language models. The results show that while its effectiveness varies by attack type and dataset, it enhances robustness and can occasionally improve overall model performance.

Our main contributions can be summarized into
three folds:
\begin{itemize}
\item We comprehensively explore the robustness of DNA language models through multi-granularity attack levels: \textbf{character-level (nucleotide-level)}, \textbf{word-level (codon-level)}, and \textbf{sentence-level (back-translation)}. Notably, we introduce \textbf{codon-level} and \textbf{back-translation} attacks, extending the scope beyond conventional character-level perturbations for a more thorough robustness evaluation.

\item We demonstrate that increasing perturbation strength causes the most severe performance degradation in nucleotide-level perturbations,thought at the risk of disrupting biological context. In the AMR gene classification with drug labels task, our proposed back-translation-based perturbations effectively degrade classification performance while preserving semantic meaning.

\item We show that even simple adversarial training effectively mitigates vulnerabilities and improves classification accuracy over the original model, highlighting its potential to enhance the real-world reliability of DNA classification models.
\end{itemize}

\section{Related Work}

\paragraph{Adversarial Attacks in Text Classification}


Adversarial attacks in text classification can be broadly categorized into character-level, word-level, and sentence-level attacks \cite{ebrahimi2018hotflip}. Character-level attacks involve random substitutions at the character level, or modifications at the word embedding or language model level \cite{jia2017adversarial, alzantot2018generating}. Word-level attacks follow a similar approach. Sentence-level attacks include paraphrasing, backtranslation \cite{ribeiro2018semantically}, and sentence embedding-level perturbations. \cite{zhang2019limitations}.


\paragraph{Adversarial Attacks in DNA Sequence Classification}


\citet{kuleshov2021quantifying} and \citet{montserrat2022adversarial} demonstrated that nucleotide-level adversarial attacks, including perturbations on single nucleotide polymorphism (SNP)-based ancestry classification models, significantly degrade the performance of DNA sequence classifiers.
In this paper, inspired by word-level and sentence-level perturbations, we further explore codon-level attacks and backtranslation-based DNA attacks.

\paragraph{Adversarial Training}
Several studies have shown that adversarial training enhances model robustness \cite{madry2018towards, Zhang2022Improving, Liu2023End}, including in DNA sequence classification, where character-level adversarial examples are used to improve resilience \cite{kuleshov2021quantifying, montserrat2022adversarial}.

\begin{figure*}
    \centering
    \includegraphics[width=1\linewidth]{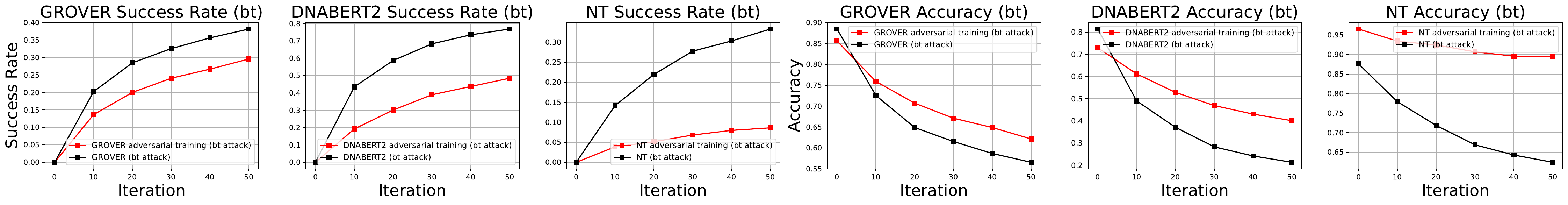}
    \caption{Comparison of success rates and accuracies between adversarial training and standard training in AMR gene Classification across increasing iterations of backtranslation attacks}
    \label{fig:sr_amr_iteration_bt}
\end{figure*}

\section{Adversarial Attack on DNA Language Models}


We examine the impact of adversarial attacks on DNABERT2 \cite{zhou2023dnabert}, Nucleotide Transformer \cite{dalla-torre2023nucleotide}, and GROVER \cite{sanabria2024grover} across three levels attack strategies: character-level (nucleotide substitutions~\cite{kuleshov2021quantifying}), word-level (codon modifications), and sentence-level (back-translation-based transformations). Our analysis focuses on DNA classification task using AMR 
 gene \cite{yoo2024predicting} and promoter detection~\cite{zhou2023dnabert}. Further implementation details and benchmark details are provided in Appendix \ref{appendix:implementation_details} and Appendix \ref{appendix:benchmark_details}.


\paragraph{Nucleotide-Level Attack}

Nucleotides are the fundamental building blocks of nucleic acids, such as DNA and RNA, crucial for storing and transmitting genetic information. In DNA sequences, nucleotides are represented by single characters, making nucleotide-level attacks analogous to character-level attacks in text. We conducted nucleotide-level attacks using an enumeration approach combined with trial and error search methods~\cite{iyyer2018adversarial, ribeiro2018semantically, belinkov2018synthetic}.

\paragraph{Codon-Level Attack}

A codon is a sequence of three nucleotides in DNA or RNA that specifies a particular amino acid. Codons are essential for protein synthesis. Codon-level attack is similar to word-level attacks and is also performed using the enumeration search method including trial and error.

\paragraph{Backtranslation Attack}

We introduce backtranslation based attack method for DNA sequences.
In biological translation, translation refers to converting mRNA sequences into protein amino acid sequences, and reverse translation refers to generating possible nucleotide sequences based on protein amino acid sequences. 

\section{Results}


\subsection{Attack Effectiveness and Context Preservation}

In Figure~\ref{fig:pd_amr_accuracy}, nucleotide-level attacks~\cite{kuleshov2021quantifying} result in the most significant performance degradation. While these attacks are highly effective, they also pose a considerable risk of altering the DNA sequence's meaning, as discussed in  Appendix~\ref{appendix:adversarial_validation}. Conversely, codon-level attacks are better at preserving sequence meaning since they group sequences based on meaningful units, such as codons. Backtranslation attacks preserve context the best but are relatively less effective in reducing accuracy. This trend is evident in the first graph, where there is a sharp decline in model accuracy with increasing epsilon values for nucleotide-level attacks, particularly affecting DNABERT2~\cite{zhou2023dnabert}. In contrast, the Nucleotide Transformer maintaines relatively high accuracy even as epsilon increased.

\subsection{Model Robustness Comparison}
The Nucleotide Transformer~\cite{dalla-torre2023nucleotide} and GROVER~\cite{sanabria2024grover} demonstrate superior robustness compared to DNABERT2~\cite{zhou2023dnabert}. Due to its larger model capacity and ability to handle longer sequences, the Nucleotide Transformer exhibits increased resistance to adversarial attacks. Compared to DNABERT2, GROVER employs an extensive BPE optimization process involving hundreds of iterations and incorporates an LSTM architecture, enhancing its ability to model long-range dependencies in DNA sequences. These architectural refinements are likely to improve its robustness against adversarial attacks. This robustness is particularly notable in experiments involving backtranslation attacks. The second graph in \autoref{fig:pd_amr_accuracy} illustrates that DNABERT2's accuracy significantly drops with increasing iterations, whereas the Nucleotide Transformer and GROVER maintain high accuracy, demonstrating its robustness due to its larger model capacity or longer sequence processing capabilities.

Moreover, DNABERT2 promoter detection model demonstrates greater robustness compared to DNABERT2 AMR drug classification models, potentially due to the larger volume of training data utilized. As illustrated in the third graph, with increasing epsilon values, the accuracy of DNABERT2 promoter detection model declines, though to a lesser extent than in AMR detection models. However, DNABERT2 still performs worse than the Nucleotide Transformer and GROVER under nucleotide-level attacks. The fourth graph further shows that, in AMR drug classification tasks, the Nucleotide Transformer and GROVER maintain the highest accuracy against backtranslation attacks, whereas DNABERT2 appears to be relatively more vulnerable to backtranslation attacks. 
We further address the robustness of general language model in Appendix~\ref{appendix:chatgpt_classification}.

\subsection{Comparative Analysis of Classification Tasks}
The experimental results indicate that the Nucleotide Transformer and GROVER possess higher robustness against adversarial attacks compared to DNABERT2, especially in the context of backtranslation attacks. Additionally, DNABERT2 trained with larger datasets, such as those used for promoter detection, demonstrates enhanced robustness. This comparative analysis highlights the importance of model capacity, training data size and The ability to process long sequences in improving the resilience of DNA sequence classification models against various adversarial attack methods.

\subsection{Performance Against Defense Method}
We further investigate adversarial training as a defense mechanism against adversarial attacks, incorporating adversarial examples into the training process to improve model robustness.

\autoref{fig:sr_amr_iteration_bt} illustrates the changes in success rate and accuracy over iterations when applying the Back-translate attack in the AMR drug classification task. The results indicate that the success rate of adversarially trained models (red) increases at a slower rate compared to standard training models (black). This suggests that adversarial training enhances the model's robustness against attacks, leading to a relatively lower attack success rate. Furthermore, the accuracy graphs show that adversarially trained models (red) exhibit a smaller decline in accuracy compared to standard training (black). This observation implies that adversarial training improves the model's generalization performance and mitigates performance degradation when exposed to adversarial attacks. For GROVER and DNABERT2, the initial accuracy of adversarially trained models is lower than that of standard training models in Figure~\ref{fig:PD_iter} and Figure~\ref{fig:AMR_iter}. However, as iterations increase, adversarially trained models demonstrate relatively higher accuracy, indicating that repeated exposure to attacks reinforce the model’s resilience. In contrast, Nucleotide Transformer exhibits superior robustness, with adversarially trained models outperforming standard training models in accuracy from the outset. Additionally, the accuracy degradation over iterations is the least pronounced among the three models. This suggests that the larger model capacity and ability to handle longer sequences of Nucleotide Transformer further amplify the benefits of adversarial training, making it more resistant to adversarial perturbations. Further details are described in Appendix~\ref{appendix:comparisons_adv_and_standard_train}

\section{Conclusion}

In this study, we utilize three methods to generate DNA adversarial examples, proposing techniques for generating adversarial examples using nucleotide based attack, codon-based attacks and backtranslation. These methods successfully degrade the performance of AMR drug classification and promoter detection models. Nucleotide-level attacks are the most effective, although they risk disrupting the DNA context. Conversely, backtranslation better preserves meaning and context. Notably, adversarial training as a defense strategy proves effective in improving model robustness, and in some cases, even surpassing the performance of standard training methods. By leveraging adversarial training, models become more resilient against such attacks.

\section*{Limitations}

This study has several limitations. First, the datasets used are limited to AMR gene classification with drug classes which may restrict the generalization of the results. Expanding experiments to include datasets such as AMR gene families or AMR mechanisms would provide a more comprehensive evaluation of adversarial robustness. 

The study evaluates adversarial attack methods on three distinct DNA language models such as DNABERT2, GROVER, and Nucleotide Transformer. While these models cover a range of architectures and sizes, extending the evaluation to additional DNA classification models could provide further insights into model-specific robustness.

Nucleotide-level attacks, though effective, risk altering the biological meaning of DNA sequences. Their impact is assessed through CG content, species recognition rate, and mapping rate via NCBI BLAST, demonstrating the effectiveness of our approach. While these metrics provide valuable insights, further validation in real biological experiments could strengthen the findings. 

Backtranslation methods help preserve sequence meaning and context, as indicated by CG content ratios and species recognition, supporting their role in enhancing model robustness. While our evaluations offer a foundation, additional experimental verification could further solidify these conclusions.


\section*{Ethical Considerations}
This study focuses on adversarial robustness in DNA language models and does not involve human subjects or sensitive data. The adversarial attack methods used are intended for evaluating and improving model reliability in bioinformatics, not for malicious purposes. However, it is important to acknowledge the potential risks of adversarial attacks in bioinformatics applications, where model vulnerabilities could impact genomic analysis and medical decision-making.



\nocite{kuleshov2021quantifying, montserrat2022adversarial,li2020bert,jin2020is, zeng2021openattack, morris2020textattack, ebrahimi2018hotflip, jia2017adversarial, alzantot2018generating, ribeiro2018semantically, zhang2019limitations, ebrahimi2018adversarial, iyyer2018adversarial, belinkov2018synthetic, madry2018towards, yoo2024predicting, zhou2023dnabert, dalla-torre2023nucleotide, alcock2020card, bonin2023megares, cook2016ebi}

\bibliography{custom}

\clearpage
\newpage
\appendix
\section{Implementation Details}
\label{appendix:implementation_details}

\begin{figure}[h]
    \centering
    \includegraphics[width=1\linewidth]{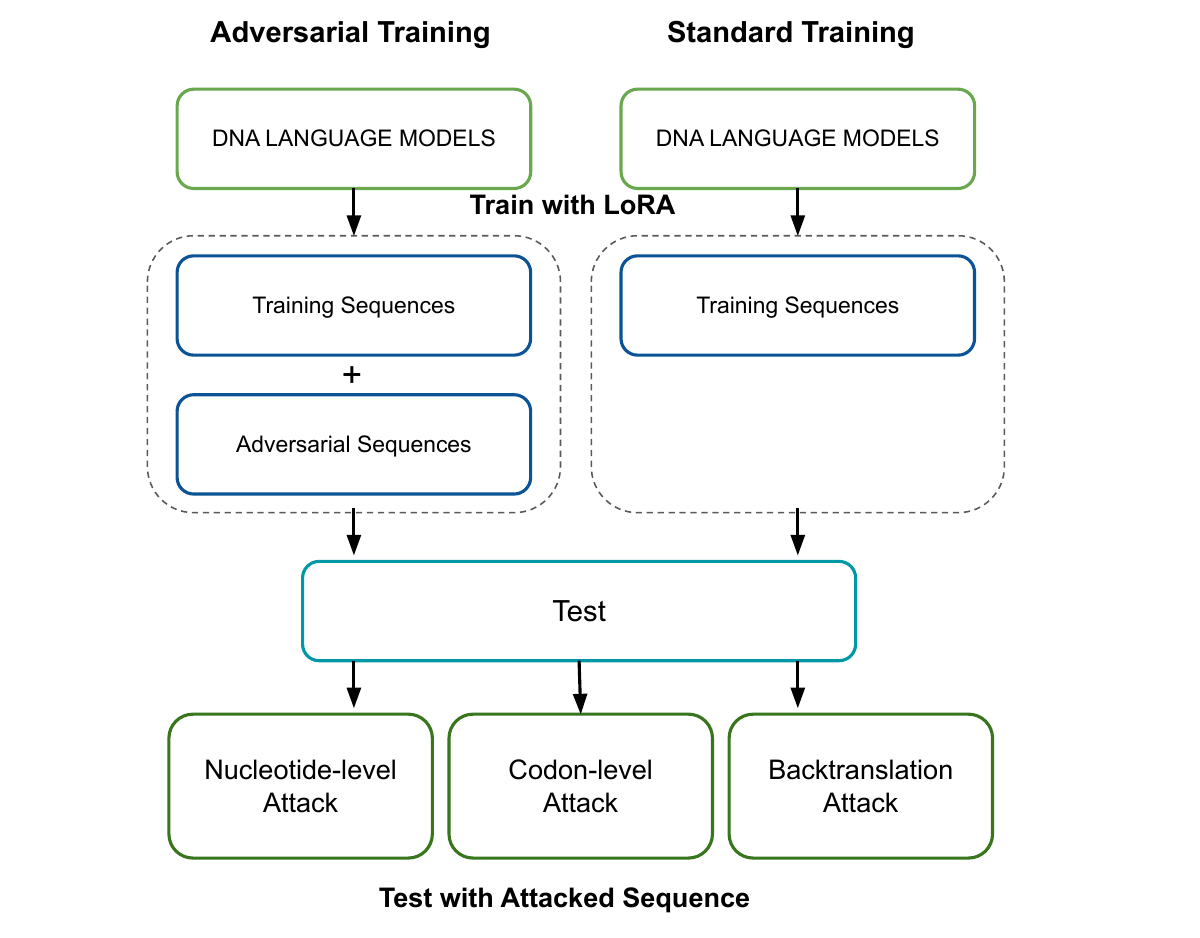}
    \caption{Overview of Adversarial Training(Fine-tuning) for DNA Language Models and Testing Adversarial Attacks}
    \label{fig:implementation_detail}
\end{figure}

We utilize LoRA \cite{hu2022lora} to finetune the DNA language models for each benchmarks. The learning rate is set to 0.0005, the number of epochs is 2, the batch size is 64. Single A100 GPU is used, and and the entire process takes approximately one hour. The number of parameters for each model is as follows: DNABERT-2 (117M), Nucleotide Transformer (2.5B), and GROVER (90M). DNABERT-2 is available under the Apache License 2.0, and Nucleotide Transformer is distributed under the CC BY-NC-SA 4.0 License. The GROVER model, as hosted on Hugging Face (PoetschLab/GROVER), does not have a publicly specified license.

\section{Details of Benchmarks}
\label{appendix:benchmark_details}
\paragraph{Antimicrobial Resistance Classification}

We employ the methodology outlined in \citet{yoo2024predicting} to integrate datasets from CARD v2 \cite{alcock2020card} and MEGARes v3 \cite{bonin2023megares} for investigating antibiotic resistance. Classes with fewer than 15 instances are excluded. The remaining data is divided into training, testing, and validation sets, maintaining the same proportions as the study. Data integration is performed using the European Bioinformatics Institute Antibiotic Resistance Ontology (EBI ARO; \citealp{cook2016ebi}), with irrelevant classes excluded. Overall, the drug class classification dataset for AMR gene classification is split into 75\% training data, 20\% test data, and 5\% validation data. There are a total of 9 drug classes. The MEGARes dataset is available under the MIT License, and the CARD dataset is distributed under the CC BY-NC 4.0 License.

\paragraph{Promoter Detection}

The promoter detection data from the Genome Understanding Evaluation~(GUE) benchmark dataset introduced in the DNABERT-2 is used to detect promoters in gene sequences \cite{zhou2023dnabert}. Promoters are DNA sequences that regulate the initiation of gene transcription and play a crucial role in gene expression regulation and understanding biological phenomena.

The data is extracted from gene sequences of various species, providing a broad context for biological research. Each data point includes both sequences that contain promoters and those that do not, enabling the model to learn how to distinguish promoters. This dataset includes complex promoter sequence elements known as MIX elements, which are found in specific genes.
We utilize the prom 300 all dataset from the GUE  benchmark for promoter detection. The dataset is splited into approximately 80\% training, 10\% development, and 10\% test data, with 2 classes for binary classification. The GUE dataset is distributed along with DNABERT-2, which is available under the Apache-2.0 License.

\section{Comparisons of Adversarial Training and Standard Training}
\label{appendix:comparisons_adv_and_standard_train}

\begin{figure*}
    \centering
    \includegraphics[width=1\linewidth]{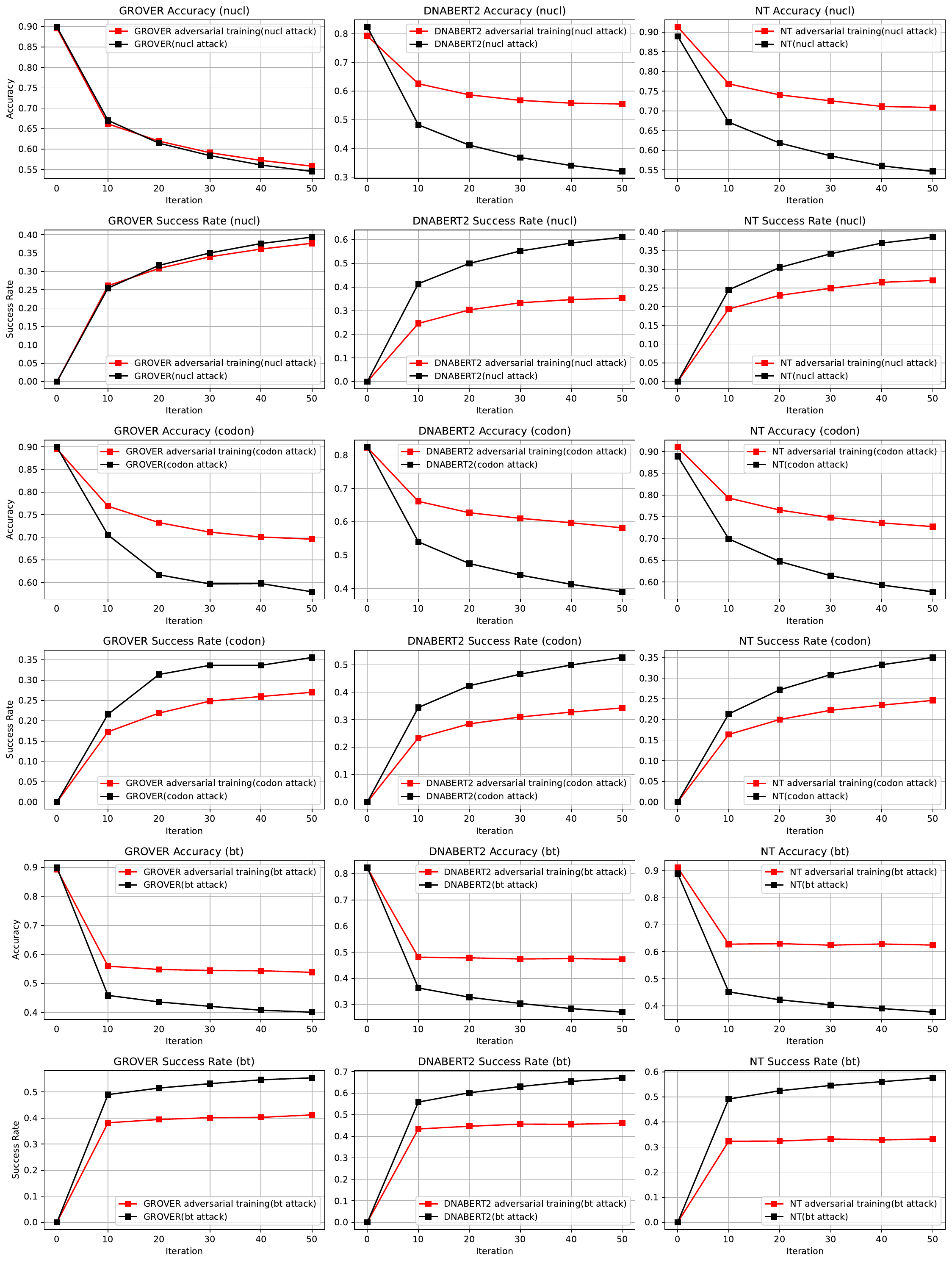}
    \caption{Comparison of Adversarial Training and Standard Training in Promoter Detection with Increasing Iterations}
    \label{fig:PD_iter}
\end{figure*}

\begin{figure*}
    \centering
    \includegraphics[width=1\linewidth]{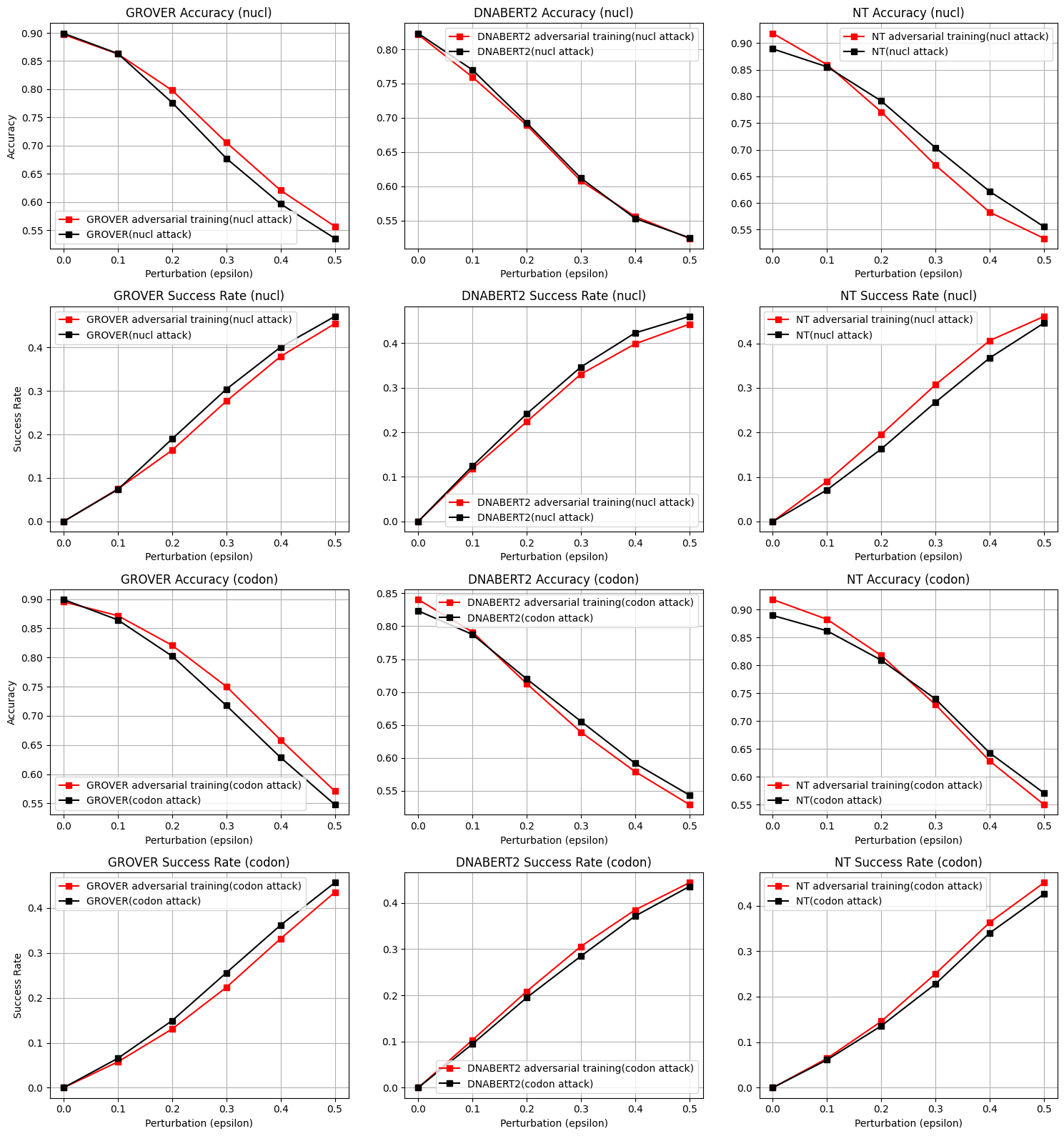}
    \caption{Comparison of Adversarial Training and Standard Training in Promoter Detection Across Different Perturbation Levels}
    \label{fig:PD_pertur}
\end{figure*}

\begin{figure*}
    \centering
    \includegraphics[width=1\linewidth]{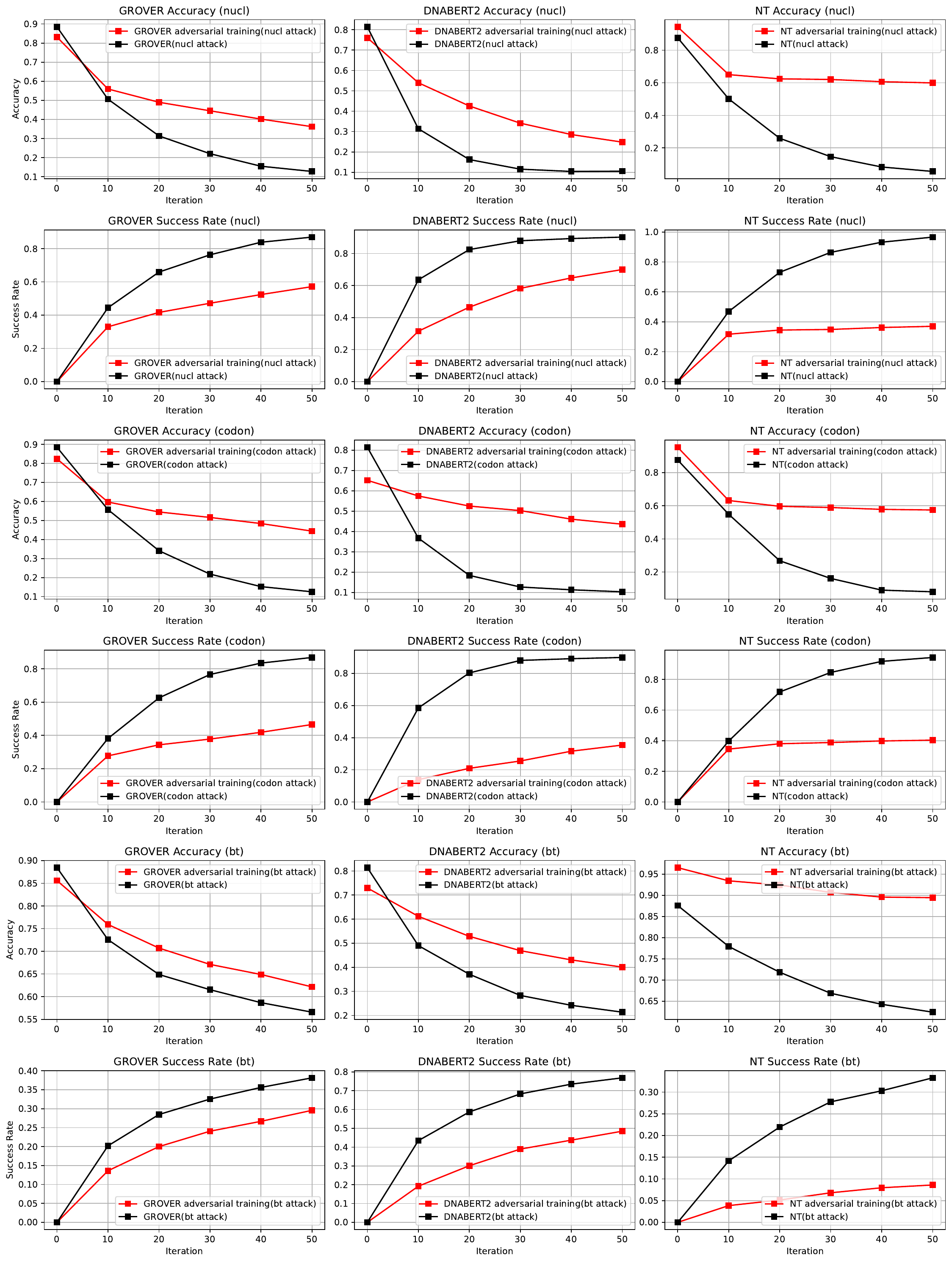}
    \caption{Comparison of Adversarial Training and Standard Training in Antimicrobial Resistance Drug Classification with Increasing Iterations}
    \label{fig:AMR_iter}
\end{figure*}

\begin{figure*}
    \centering
    \includegraphics[width=1\linewidth]{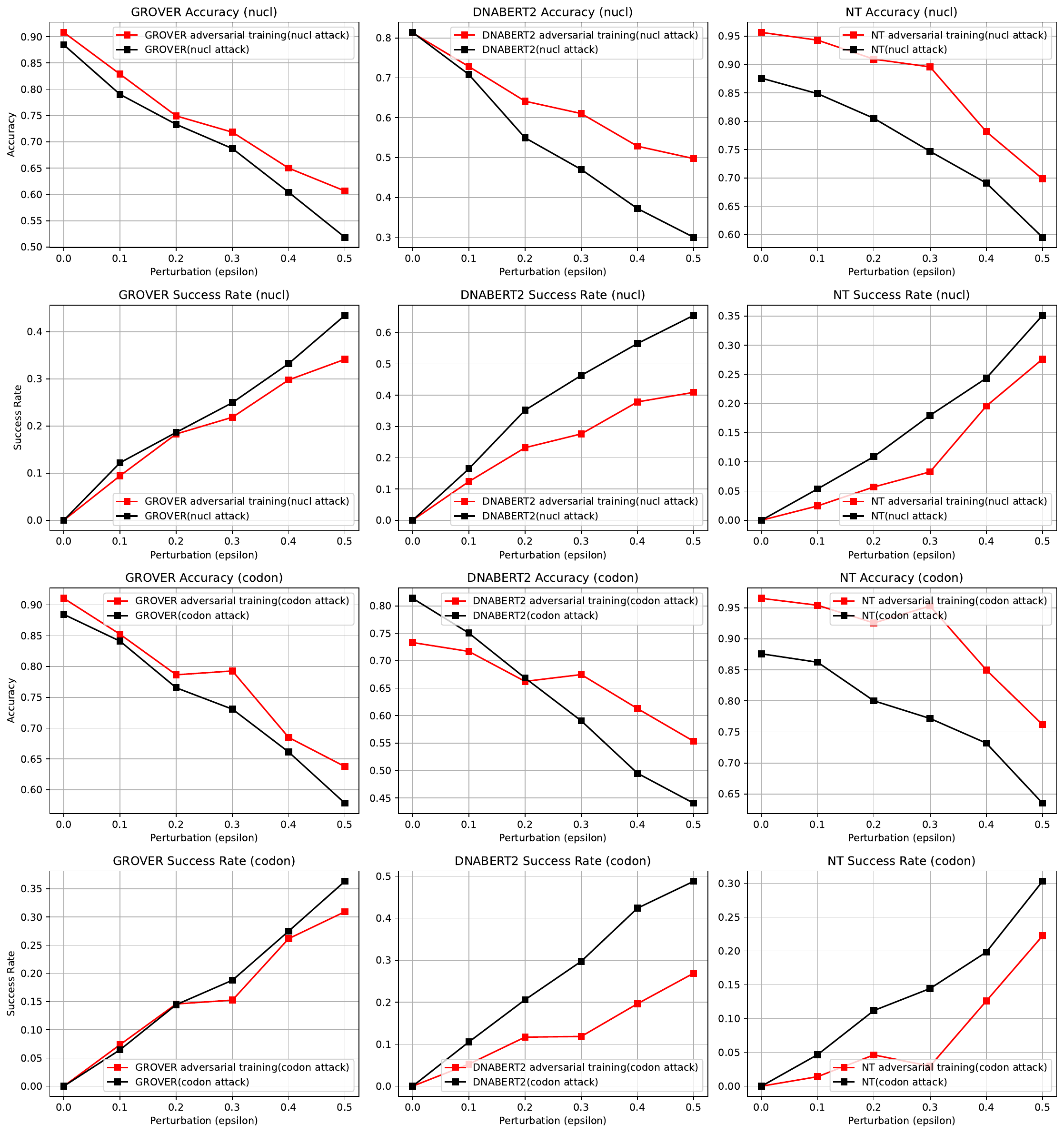}
    \caption{Comparison of Adversarial Training and Standard Training in Antimicrobial Resistance Drug Classification Across Different Perturbation Levels}
    \label{fig:AMR_pertur}
\end{figure*}

\autoref{fig:PD_iter}, \autoref{fig:PD_pertur}, \autoref{fig:AMR_iter}, and \autoref{fig:AMR_pertur} compare the performance differences between adversarially trained models and conventionally trained models (black) as the number of attack iterations and perturbation magnitude (epsilon) increase. Epsilon represents the rate of change in the sequence for the attack, indicating the proportion of the sequence that is altered. Iterations indicate the number of times the attack is performed. While all models experience a decrease in accuracy as the number of iterations and perturbation levels increase, adversarially trained models generally maintain higher accuracy and exhibit a more gradual decline in performance. In terms of attack success rate, conventionally trained models show a sharp increase in attack success as the number of iterations and perturbation magnitude grow, whereas adversarially trained models exhibit a more moderate increase, demonstrating greater resistance to adversarial attacks. These findings suggest that adversarial training effectively enhances model robustness against iterative attacks and higher perturbation magnitudes while mitigating performance degradation. Figures present accuracy and success rate trends under different perturbation settings. However, all results are based on a single run.

\section{Adversarial DNA Sequence validation}
\label{appendix:adversarial_validation}

\subsection{Promoter Detection Sequence validation}

\begin{equation}
GC\% = \frac{G+C}{A+T+G+C} \times 100
\label{eq:gc_content}
\end{equation}

To validate adversarial sequences in promoter detection, we compared the GC content of the original sequences and the adversarially generated sequences. Prior research has highlighted the strong correlation between promoter regions and GC content \cite{Umarov2019}. Using this insight, we assess whether the adversarial sequences maintain a similar GC content distribution to the original sequences. We calculate the GC content percentage using Equation~\ref{eq:gc_content}.

Specifically, we measure the Pearson correlation coefficient \cite{pearson1896regression} between the GC content of original and adversarial sequences to quantify their similarity. This approach allows us to evaluate whether the adversarial sequences retain key GC content characteristics, providing an initial validation measure.

\begin{table*}[h]
  \centering
  \begin{tabular}{ccc}
    \hline
    \textbf{$\epsilon$} & \textbf{Nucleotide-level Attack} & \textbf{Codon-level Attack} \\
    \hline
    0.1 & 0.9910 & 0.9903 \\
    0.2 & 0.9788 & 0.9776  \\
    0.3 & 0.9614 & 0.9592  \\
    0.4 &  0.9361 & 0.9327   \\
    0.5 & 0.8990 &  0.8970 \\
    \hline
  \end{tabular}
  \caption{Pearson Correlation Coefficient of GC content based on perturbation ($\epsilon$) for different attack strategies.}
  \label{tab:PD_perturbation_comparison}
\end{table*}

\begin{table*}[h]
  \centering
  \begin{tabular}{cccc}
    \hline
    \textbf{Iteration} & \textbf{Nucleotide-level Attack} & \textbf{Codon-level Attack} & \textbf{Back-translation Attack} \\
    \hline
    10 & 0.9525 & 0.9481 &  0.9303 \\
    20 & 0.8911 & 0.8925 & 0.9271 \\
    30 & 0.8512 & 0.8498 & 0.9264\\
    40 &  0.8236 & 0.8193 &  0.9250 \\
    50 & 0.8068 & 0.7987  &  0.9270 \\
    \hline
  \end{tabular}
  \caption{ Pearson Correlation Coefficient of GC content based on iteration for different attack strategies.}
  \label{tab:PD_iteration_comparison}
\end{table*}

Table \ref{tab:PD_perturbation_comparison} demonstrates a decreasing trend in the Pearson correlation coefficient of GC content between the original and perturbed sequences as the perturbation level increases. This trend is observed in both nucleotide-level and codon-level attacks, where a higher perturbation parameter (\(\epsilon\)) results in a lower Pearson correlation coefficient.
Table \ref{tab:PD_iteration_comparison} presents the Pearson correlation coefficient between the original sequences and perturbed sequences as the number of iterations increases for nucleotide-level, codon-level, and backtranslation-based attacks. Across all attack methods, the Pearson correlation coefficient decreases as the number of iterations increases. However, in the case of backtranslation, the decline is less pronounced compared to nucleotide-level and codon-level attacks, which may suggest that backtranslation better preserves the contextual integrity of the original sequences.

\subsection{Antimicrobial Resistance Drug classification Sequence validation}

\begin{table*}[h]
  \centering
  \begin{tabular}{ccc}
    \hline
    \textbf{$\epsilon$} & \textbf{Nucleotide-level Attack} & \textbf{Codon-level Attack} \\
    \hline
    0.1 & 93.05 & 92.60 \\
    0.2 & 70.35 & 85.98  \\
    0.3 & 16.25 & 75.81  \\
    0.4 &  1.36 & 46.15   \\
    0.5 & 0.50 & 13.03  \\
    \hline
  \end{tabular}
  \caption{Species Matching rate (\%) based on perturbation ($\epsilon$) for different attack strategies.}
  \label{tab:AMR_perturbation_comparison_mapping}
\end{table*}

\begin{table*}[h]
  \centering
  \begin{tabular}{ccc}
    \hline
    \textbf{$\epsilon$} & \textbf{Nucleotide-level Attack} & \textbf{Codon-level Attack} \\
    \hline
    0.1 & 99.75 & 99.88 \\
    0.2 & 84.49 & 99.88  \\
    0.3 & 25.56 & 96.40  \\
    0.4 &  2.61 & 68.98   \\
    0.5 & 0.62 & 24.81  \\
    \hline
  \end{tabular}
  \caption{Recognition Rate (\%) based on perturbation ($\epsilon$) for different attack strategies.}
  \label{tab:AMR_perturbation_comparison_recognition}
\end{table*}

\begin{table*}[h]
  \centering
  \begin{tabular}{cccc}
    \hline
    \textbf{Iteration} & \textbf{Nucleotide-level Attack} & \textbf{Codon-level Attack} & \textbf{Back-translation Attack} \\
    \hline
    10 & 0.25 & 1.24 & 1.99 \\
    20 & 0.25 & 0.25 & 2.73 \\
    30 & 0.25 & 0.25 & 3.47 \\
    40 &  0.25 & 0.25 & 1.61 \\
    50 & 0.25 & 0.25 & 3.10 \\
    \hline
  \end{tabular}
  \caption{Species Matching rate (\%) based on iteration for different attack strategies.}
  \label{tab:AMR_iteration_comparison_mapping}
\end{table*}

\begin{table*}[h]
  \centering
  \begin{tabular}{cccc}
    \hline
    \textbf{Iteration} & \textbf{Nucleotide-level Attack} & \textbf{Codon-level Attack} & \textbf{Back-translation Attack} \\
    \hline
    10 & 0.25 & 1.99 & 15.63 \\
    20 & 0.25 & 0.25 & 16.63 \\
    30 & 0.25 & 0.25 & 18.11 \\
    40 &  0.25 & 0.25 & 18.73 \\
    50 & 0.37 & 0.25  & 16.63 \\
    \hline
  \end{tabular}
  \caption{Species Recognition rate (\%) based on iteration for different attack strategies.}
  \label{tab:AMR_iteration_comparison_recognition}
\end{table*}

To evaluate the effectiveness of adversarial DNA sequences for attacks as the number of iterations and the magnitude of perturbations increase, we utilize the BLAST (Basic Local Alignment Search Tool\cite{altschul1990blast}) database mapping API provided by the National Center for Biotechnology Information (NCBI)\cite{Johnson2008}.

Analysis of the BLAST mapping results reveal a gradual decline in the mapping rate of adversarially perturbed sequences as the same species as the original test dataset, as the intensity of adversarial perturbations increased. Specifically, when the perturbation magnitude is small, the perturbed sequences are mapped as the same species as the original sequence. However, as the perturbation magnitude increases, the likelihood of being classified as a different species increases. This indicates that adversarial attacks effectively induce classification perturbations in the intended direction. However, if the perturbed sequences are classified as entirely different species, the effectiveness of the attack sequences may be somewhat diminished. Ideally, a adversarial sequence should not excessively distorted the biological properties of the sequence. Excessive perturbation, however, may result in adversarial sequences being recognized as entirely different species, which could undermine their intended effectiveness.

Additionally, we examine cases where BLAST fails to clearly match the input sequence to any species in the existing database (i.e., cases where BLAST does not recognize the sequence). The results indicate that as the intensity of the adversarial attack increases, the proportion of sequences that BLAST fails to recognize rose sharply. This suggests that larger perturbations lead to greater distortion of the biological properties of the original DNA sequences. In particular, for iterations, both the mapping rate and recognition rate are significantly low, suggesting that the results may not fully align with real-case scenarios. However, similar to the validation of promoter detection sequences, applying backtranslation results in higher mapping and recognition rates compared to nucleotide-level and codon-level attacks. This may indicate that backtranslation relatively better preserves the contextual integrity of the sequences.
Consequently, in such cases, the effectiveness of the adversarial sequences may also be reduced, further distancing them from the natural noise present in real-world conditions. This is particularly evident in multi-iteration nucleotide and codon-level attacks, as well as backtranslation, where recognition and mapping rates in the BLAST database are significantly low, making them a form of worst-case performance exploration.

\section{Adversarial Evaluation of GPT-4o on DNA AMR Classification}
\label{appendix:chatgpt_classification}

\begin{figure}
    \centering
    \includegraphics[width=1\linewidth]{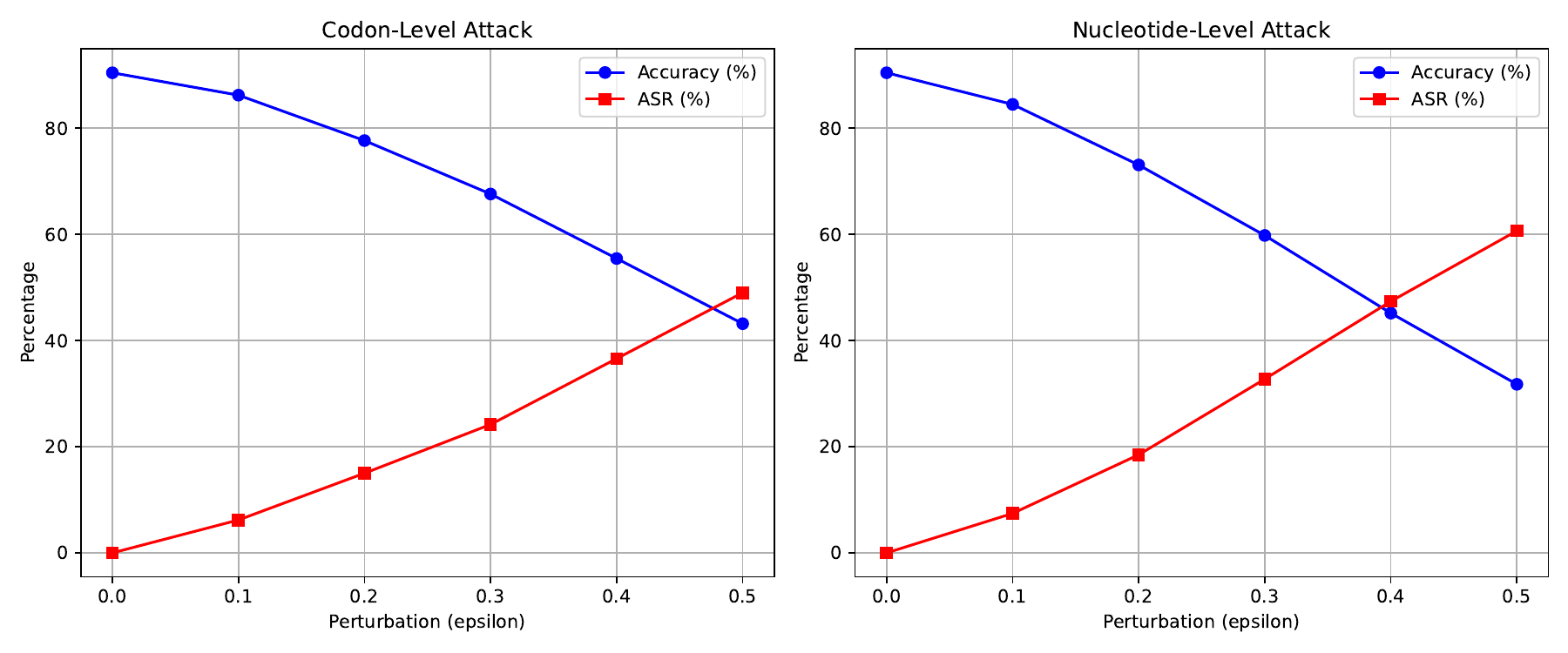}
    \caption{Accuracy and attack success rate (ASR) under codon-level and nucleotide-level perturbations in Antimicrobial Resistance Drug Classification Across Different Perturbation Levels}
    \label{fig:chatgpt4mini_attack_results}
\end{figure}

We present additional analysis on the adversarial robustness of \texttt{gpt-4o-mini-2024-07-18} fine-tuned for DNA AMR classification. This study has demonstrated that DNA Language Models are vulnerable to adversarial attacks, and we extend this investigation to examine whether general-purpose LLMs, when adapted for biological sequence classification, exhibit similar weaknesses.

By applying codon-level and nucleotide-level perturbations, we observe a sharp decline in model accuracy as perturbation intensity increased (dropping to 31.76\% at the highest level), while the attack success rate (ASR) rises to a maximum of 60.67\% in Figure~\ref{fig:chatgpt4mini_attack_results}.

With the growing interest in applying LLMs to DNA sequence analysis~\cite{Sarumi2024LLMBioinformatics,Liu2024LLMBioinformatics}, ensuring robustness remains a critical challenge. Adversarial training and domain-specific robustness strategies may be necessary to enhance the reliability of such models in biological sequence tasks.

\section{Effect of Realistic Sequencing Errors on Robustness}
To evaluate model robustness under more realistic sequencing scenarios, we simulate sequencing errors using PBSIM2~\cite{Ono2021PBSIM2}, incorporating real-world error patterns observed in PacBio and Oxford Nanopore sequencing. Sequencing Error Simulation is a technique used to model errors occurring during the genome sequencing process, playing a crucial role in genome analysis, algorithm evaluation, and robustness assessment in bioinformatics. PBSIM2, widely used for this purpose, simulates insertion, deletion, and substitution errors, as well as the length distribution observed in long-read sequencing technologies such as PacBio and Oxford Nanopore, enabling the generation of reads that closely resemble real sequencing data. The simulated sequencing error data can be utilized to test model robustness by incorporating noise similar to that encountered in real sequencing experiments. This holds significant importance from the perspective of adversarial training and robustness assessment, as it helps deep learning-based genomic analysis models maintain predictive accuracy while improving their resilience to sequencing errors. Our experiments show that the model's performance degrades under these conditions, indicating that realistic sequencing errors can pose a challenge to robustness. In the promoter detection task with Nucleotide Transformer, accuracy dropped from 0.8892 to 0.8608.

\end{document}